\documentclass[11pt,a4paper]{article}
\usepackage[hyperref]{acl2018}
\usepackage{times}
\usepackage{latexsym}
\usepackage{graphicx}  
\usepackage{subfigure}
\usepackage{tabularx}
\usepackage{array}
\usepackage{amsmath}
\usepackage{amssymb}
\usepackage{mathrsfs}
\usepackage{mathtools}
\usepackage{bm}
\usepackage{caption}
\usepackage[flushleft]{threeparttable}
\usepackage{url}

\aclfinalcopy 
\def\aclpaperid{574} 

\title{Dynamic Sentence Sampling for \\ Efficient Training of Neural Machine Translation}

\author{Rui Wang, Masao Utiyama, and Eiichiro Sumita \\
	National Institute of Information and Communications Technology (NICT) \\
	3-5 Hikari-dai, Seika-cho, Soraku-gun, Kyoto, Japan \\
	{\tt \{wangrui, mutiyama, eiichiro.sumita\}@nict.go.jp}}

\date{}

\begin{document}
\maketitle
\begin{abstract}
  Traditional Neural machine translation (NMT)  involves a fixed training procedure where each sentence is sampled once during each epoch. In reality, some sentences are well-learned during the initial few epochs; however, using this approach, the well-learned sentences would continue to be trained along with those sentences that were not well learned for 10-30 epochs, which results in a wastage of time. Here, we propose an efficient method to dynamically sample the sentences in order to accelerate the NMT training. In this approach, a weight is assigned to each sentence based on the measured difference between the training costs of two iterations. Further, in each epoch, a certain percentage of sentences are dynamically sampled according to their weights. Empirical results based on the NIST Chinese-to-English and the WMT English-to-German tasks show that the proposed method can significantly accelerate the NMT training and improve the NMT performance.
\end{abstract}

\section{Introduction}
\label{sec:intr}

Recently neural machine translation (NMT) has been prominently used to perform various translation tasks \citep{luong2015stanford,bojar-EtAl:2017:WMT1}. However, NMT is much more time-consuming than traditional phrase-based statistical machine translation (PBSMT) due to its deep neural network structure. To improve the efficiency of NMT training, most of the studies focus on reducing the number of parameters in the model \citep{DBLP:journals/corr/SeeLM16,DBLP:journals/corr/CregoKKRYSABCDE16,DBLP:journals/corr/HubaraCSEB16} and implementing parallelism in the data or in the model \citep{DBLP:journals/corr/WuSCLNMKCGMKSJL16,DBLP:journals/corr/KalchbrennerESO16,DBLP:journals/corr/GehringAGYD17,DBLP:journals/corr/VaswaniSPUJGKP17}.

Although these technologies have been adopted, deep networks have to be improved to achieve state-of-the-art performance in order to handle very large datasets and several training iterations. Therefore, some researchers have proposed to accelerate the NMT training by resampling a smaller subset of the data that makes a relatively high contribution, to improve the training efficiency of NMT. Specifically,  \citet{DBLP:journals/corr/KocmiB17aa} empirically investigated curriculum learning based on the sentence length and word rank. \citet{wang-EtAl:2017:Short3} proposed a static sentence-selection method for domain adaptation using the internal sentence embedding of NMT. They also proposed a sentence weighting method with dynamic weight adjustment \cite{wang-EtAl:2017:EMNLP20174}. \citet{vanderwees-bisazza-monz:2017:EMNLP2017} used domain-based cross-entropy as a criterion to  gradually fine-tune the NMT training in a dynamical manner. All of these criteria \citep{wang-EtAl:2017:Short3,wang-EtAl:2017:EMNLP20174,vanderwees-bisazza-monz:2017:EMNLP2017} are calculated before performing the NMT training based on the domain information and are fixed while performing the complete procedure. \citet{zhang-EtAl:2017:I17-2} adopted the sentence-level training cost as a dynamic criterion to gradually fine-tune the NMT training. This approach was developed based on the idea that the training cost is a useful measure to determine the translation quality of a sentence.  However, some of the sentences that can be potentially improved by training may be deleted using this method. In addition, all of the above works primarily focused on NMT translation performance, instead of training efficiency.

In this study, we propose a method of dynamic sentence sampling (DSS) to improve the NMT training efficiency. First, the differences between the training costs of two iterations, which is a measure of  whether the translation quality of a sentence can be potentially improved, is measured to be the criterion.  We further proposed two sentence resampling strategies, i.e.,  weighted sampling and  review mechanism to help NMT focus on the not well-learned sentences as well as remember the knowledge from the well-learned sentences.

The remainder of this paper is organized as follows. In Section \ref{sec:edss}, we introduce the dynamic sentence sampling method.  Experiments are described and analyzed in Section \ref{sec:ex}. We discussed some other effects of the proposed methods in Section \ref{sec:dis}. We conclude our paper in the last section.

\section{Dynamic Sentence Sampling (DSS)}
\label{sec:edss}
\subsection{NMT Background}
\label{sec:back}
An attention-based NMT system uses a bidirectional RNN as an encoder and a decoder that emulates the search through a source sentence during the decoding process
\citep{DBLP:journals/corr/BahdanauCB14,luong-pham-manning:2015:EMNLP}. The training objective function to be minimized can be formulated as:

	\begin{equation}
		\label{eq:oj}
		\begin{aligned}
			J = \sum_{\langle x, y\rangle \in D} - \log P(y|x, \bm{\theta}),
		\end{aligned}
	\end{equation}
where $\langle x, y\rangle$ is the parallel sentence pair from the training corpus $D$,  $P(y|x)$ is the translation probability, and $\bm{\theta}$ is the neural network parameters.

\subsection{Criteria}
\label{sec:cr}
The key to perform sentence sampling is to measure the criteria. As we know, the NMT system continually alters throughout the training procedure. However, most of the criteria described in the introduction remain constant during the NMT training process. \citet{zhang-EtAl:2017:I17-2} adopted the  sentence-level training cost to be a dynamic criterion; further, the training cost of a sentence pair $\langle x, y\rangle$ during the $i$th iteration can be calculated as:

	\begin{equation}
		\label{eq:cost}
		\begin{aligned}
			cost_{{\langle x, y\rangle}}^i= -\log P(y|x, \bm{\theta}).
		\end{aligned}
	\end{equation}

Directly adopting training cost as the criterion to select the top-ranked sentences that represent the largest training costs has two drawbacks: 1) The translation qualities of sentences with small training costs may be further improved during the succeeding epochs. 2) If the training corpus become smaller after each iteration, the knowledge associated with the removed sentences may be lost over the course of the NMT process. 

Therefore, we adopt the ratio of differences ($dif$) between training costs of two training iterations to be the criterion,

	\begin{equation}
		\label{eq:cost-diff}
		\begin{aligned}
			dif_{{\langle x, y\rangle}}^i= \frac{cost_{{\langle x, y\rangle}}^{i-1} - cost_{{\langle x, y\rangle}}^i} {cost_{{\langle x, y\rangle}}^{i-1}}. \\
		\end{aligned}
	\end{equation}

It should be noted that some of $dif_{{\langle x, y\rangle}}$ are negative. That is, the costs of some sentence pairs even increase after one epoch training. Therefore, the difference is normalized into [0, 1] as the final criterion: 

\begin{equation}
	\label{eq:cost-diff-norm}
	\begin{aligned}
		criterion_{{\langle x, y\rangle}}^i = \frac{dif_{{\langle x, y\rangle}}^i - min(dif^i)} {max(dif^i) - min(dif^i)}.
	\end{aligned}
\end{equation}

This criterion indicates the likelihood of a sentence to be further improved in the next iteration; low values indicate that the training cost of a sentence is unlikely to change and that it would not significantly contribute to the NMT training even if the sentence was trained further.

\subsection{Dynamic Sampling}
\label{sec:ds}
As we know, the NMT performance improves significantly during the initial several epochs and less significantly thereafter. This is partially because that some of the sentences have been learned sufficiently (i.e., low  $criterion_{{\langle x, y\rangle}}^i$ values). However, they are kept further training with the ones which have not been learned enough (i.e., high  $criterion_{{\langle x, y\rangle}}^i$ values). Therefore, in this approach, these sentences are deleted for the subsequent iterations. To ensure that knowledge from the deleted sentences is retained, we propose two mechanisms for dynamic sampling, which are described in the succeeding sections.

\subsubsection{Weighted Sampling (WS)}
\label{sec:ws}

We assign a normalized weight to each sentence according to the criterion that can be given as:

	\begin{equation}
		\label{eq:ws}
		\begin{aligned}
			weight_{{\langle x, y\rangle}}^i = \frac{criterion_{{\langle x, y\rangle}}^i } {\sum_{\langle x, y\rangle \in D} criterion_{{\langle x, y\rangle}}^i}.
		\end{aligned}
	\end{equation}

Further, weighted sampling without any replacement was used to select a small subset, such as 80\%\footnote{\citet{zhang-EtAl:2017:I17-2} adopted 80\% as the selection threshold and we follow their settings for fair comparison. Due to limited space, we will empirically investigate the effect of the thresholds as our future work. } of the entire corpus, as the corpus $D_{ws}^{i+1}$ to perform the subsequent iteration. The updated objective function using weighted sampling $J_{ws}$ can be formulated as follows:

	\begin{equation}
		\label{eq:wsj}
		\begin{aligned}
			J_{ws}=\sum_{\langle x, y\rangle \in D_{ws}} -\log P(y|x, \bm{\theta}).
		\end{aligned}
	\end{equation}

Thus only 80\% of the entire corpus is used to perform the NMT training during each iteration (for the first two iteration, all of the sentences should be sampled). Because the criterion continually changes, the sentence selection procedure also changes during the NMT training. Those that are not selected in an epoch still have a chance to be selected in the subsequent epoch\footnote{For those 20\% sentences who are not selected, their $criterion_{{\langle x, y\rangle}}^{i+1} =  criterion_{{\langle x, y\rangle}}^{i}$.}.  

\begin{figure*}[ht]
	\centering
	\includegraphics[width=6.3in]{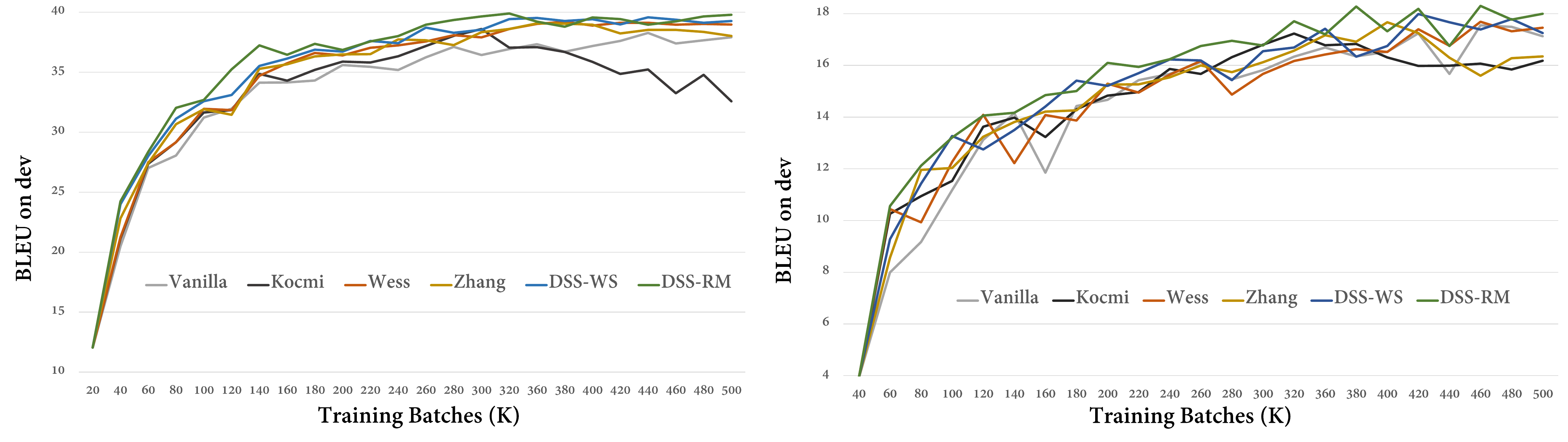}		
	\caption{Learning curves. Left: NIST ZH-to-EN; Right EN-to-DE.}\label{fig:curve}
\end{figure*}

\subsubsection{Review Mechanism (RM)}
\label{sec:rm}

We further propose an alternate sentence sampling mechanism. After performing an iteration during training, 80\% of the top-ranked  sentences are selected to act as the training data for the subsequent iteration. Each sentence that is not selected is classified into the low-criterion group $D_{low}$ and does not have a chance to be sampled again. In this case, the $D_{low}$ will become larger and larger, and $D_{high}$ will becomes smaller and smaller. To prevent the loss of the knowledge that  was obtained from the $D_{low}$ group during NMT, a small percentage $\lambda$, such as 10\%, of the  $D_{low}$ group is sampled as the knowledge to be reviewed. The updated NMT objective function is formalized as follows,

{\footnotesize 
	\begin{equation}
		\label{eq:rm}
		\begin{aligned}
			J_{rm} = \sum_{\mathclap{\langle x, y\rangle \in D_{high}}} -\log P(y|x, \bm{\theta}) + \sum_{\mathclap{\langle x, y\rangle \in \lambda  D_{low}}} - \log P(y|x, \bm{\theta}).
		\end{aligned}
	\end{equation}}

\section{Experiments}
\label{sec:ex}

\subsection{Datasets}

The proposed methods were applied to perform 1) the NIST Chinese (ZH) to English (EN) translation task that contained a training dataset of 1.42 million bilingual sentence pairs from LDC corpora\footnote{LDC2002E18, LDC2003E07, LDC2003E14, Hansards portion of LDC2004T07, LDC2004T08, and LDC2005T06.}. The NIST02 and NIST03-08 datasets were used as the development and test datasets, respectively. 2) the WMT  English to German (DE) translation task for which 4.43 million bilingual sentence pairs from the WMT-14 dataset\footnote{\url{https://nlp.stanford.edu/projects/nmt/data/wmt14.en-de/}}  was used as the training data. The newstest2012 and newstest2013-2015 datasets were used as development and test datasets, respectively.

\begin{table*}[ht]\small
	\caption{\label{tab:zh}Results from the NIST ZH-to-EN translation task.}
	\begin{center}
		\begin{tabular}{l|l|l|l|l|l|l|l}
			\hline
			\hline
			Methods & Dev (NIST02) & NIST03 & NIST04 &NIST05&NIST06&NIST08& Test (all)\\
			\hline
			PBSMT&33.15 &  31.02 & 33.78 & 30.33 &29.62&23.53& 29.66\\
			\hline
			Vanilla NMT &  38.48 & 37.53 & 39.95 &35.24&33.86&27.23& 35.08\\
			Random Sampling &38.35  &36.45 &  40.01&34.27&33.70&26.37& 34.62\\
			\citet{DBLP:journals/corr/KocmiB17aa}   &  38.51& 37.60& 39.87 &35.43&33.76&27.37& 35.19\\
			\citet{vanderwees-bisazza-monz:2017:EMNLP2017}  & \textbf{39.16}&  38.09&  40.30&\textbf{35.59}&\textbf{34.14}&27.46& \textbf{35.62}\\
			\citet{zhang-EtAl:2017:I17-2}  & 39.08 &  \textbf{38.27}& \textbf{40.37} &35.32&33.57&\textbf{27.87}& 35.57\\
			\hline
			DSS-WS & 39.54+&  39.23++&  \textbf{40.84}+&\textbf{35.98}+&34.91++&28.42+&36.85++ \\
			DSS-RM &   \textbf{39.89}++&  \textbf{39.90}++&  40.60&35.77+&\textbf{35.45}++&\textbf{29.30}++& \textbf{37.33}++\\
			\hline
			\hline
		\end{tabular}
	\end{center}
	\caption{\label{tab:de}Results from the WMT EN-to-DE translation task.}
	\begin{center}
		\begin{tabular}{l|l|l|l|l|l}
			\hline
			\hline
			Methods &  Dev (newstest2012)& newstest2013 & newstest2014 &newstest2015& Test (all)\\
			\hline
			PBSMT & 14.89& 16.75& 15.19 &16.84& 16.35\\
			\hline
			Vanilla NMT &  17.55& 20.92 & 19.16 &20.01& 20.06\\
			Random Sampling & 17.39& 20.32 & 18.36 &20.30& 19.61\\
			\citet{DBLP:journals/corr/KocmiB17aa} &  17.63& 20.63 & 19.21 &\textbf{20.47}& 20.18\\
			\citet{vanderwees-bisazza-monz:2017:EMNLP2017} & \textbf{17.69}& \textbf{20.81} & 19.21 &20.24& 20.19\\
			\citet{zhang-EtAl:2017:I17-2} &17.67 & 20.80 & \textbf{19.37} &20.42& \textbf{20.30}\\
			\hline
			DSS-WS & 17.99& 21.11 & 19.89+ &\textbf{21.20}+& 20.96+\\
			DSS-RM &  \textbf{18.34}+& \textbf{21.76}++ & \textbf{20.04}++ &21.02+& \textbf{21.22}++\\
			\hline
			\hline
		\end{tabular}
	\end{center}
	\begin{tablenotes}
		\item Note: The translation performance was measured using the case-insensitive BLEU \citep{Papineni:2002:BMA:1073083.1073135} scores. Marks after the scores indicate that the proposed methods significantly \citep{koehn:2004} outperformed the existing optimal baselines in bold (\textquotedblleft $++$\textquotedblright denotes better at a significance  level of $\alpha = 0.01$, whereas \textquotedblleft $+$\textquotedblright denotes better at a significance level of $\alpha = 0.05$.).
	\end{tablenotes}

\end{table*}

\subsection{Baselines and Settings}
Beside the  PBSMT \citep{koehn-EtAl:2007:PosterDemo} and vanilla NMT, three typical existing approaches described in the introduction were empirically compared: 1)   Curriculum learning using the source sentence length as the criterion \citep{DBLP:journals/corr/KocmiB17aa}. 2)   Gradual fine-tuning using language model-based cross-entropy  \citep{vanderwees-bisazza-monz:2017:EMNLP2017}\footnote{\citet{vanderwees-bisazza-monz:2017:EMNLP2017} also proposed a weighted sampling method; however, its performance was worse than that of the gradual fine-tuning. The method originally adopted by Wees et al. was based on the cross-entropy differences between two domains. Because no domain information is available for this task; the development data was used as the in-domain data by that method. In the method proposed in this study, the development data is not required.}.  3) NMT boosting method by eliminating 20\% of the training data with the lowest training cost after performing every iteration \citep{zhang-EtAl:2017:I17-2}.

For the proposed DSS method, we adopted one epoch as one iteration for the EN-DE task and three epochs as one iteration for the ZH-EN task, because the corpus size of the EN-DE task is approximately three times larger than that of  the ZH-EN task.

\subsection{NMT Systems}
The proposed method was implemented in Nematus \citep{sennrich2017nematus} with the following default settings:  the word embedding dimension was 620, the size of each hidden layer was 1,000, the batch size was 80, the maximum sequence length was 50, and the beam size for the decoding was 10.   A 30K-word vocabulary was created and data was shuffled before each epoch. Training was conducted on a single Tesla P100 GPU using default dropout and  the ADADELTA optimizer \citep{zeiler2012adadelta} with default learning rate 0.0001. All of the systems were trained for 500K batches which took approximately 7 days.

\subsection{Results and Analyses}

\subsubsection{Training Efficiency}
The learning curve is depicted in Figure \ref{fig:curve}.  

1) The BLEU score (ZH-EN as example) of vanilla NMT increased from 0 to 35 using approximately 200K training batches.  Further, the BLEU increased from 35 to 38 using around 200K additional training batches. This is consistent with our hypothesis that the improvement in NMT shows decreasing significance as the training progresses.

2) For the baselines, the method developed by \citet{DBLP:journals/corr/KocmiB17aa} did not provide significant improvement in speed. The method proposed by \citet{vanderwees-bisazza-monz:2017:EMNLP2017} and \citet{zhang-EtAl:2017:I17-2} slightly accelerated the NMT training. 

3) The proposed DSS methods significantly accelerated the NMT training. The BLEU score (ZH-EN as example) reached 35 after using approximately 140K training batches; further, the BLEU score reached 38 after using approximately additional 120K training batches. This may be caused due to the fact that the amount of well-learned became larger and larger as the training kept going. If these sentences were continually trained, the performance would not increase significantly. In comparison, DSS methods eliminated these well-learned sentences; therefore, the performance kept improving significantly until all of the sentences become well-learned. 

4) The performances of \citet{DBLP:journals/corr/KocmiB17aa} and \citet{zhang-EtAl:2017:I17-2} decreased significantly after reaching the highest BLEU. This is consistent with the hypothesis that NMT may forget  the learned knowledge by directly removing corresponding sentences. In comparison, the performances of the proposed DSS methods did not decrease significantly, because the removed sentences still have chances to be sampled.

\subsubsection{Translation Performance}
For fair comparison, we evaluated the best performed (on dev data) model during 500K training  batches on the test data. The results are shown in Tables \ref{tab:zh} and \ref{tab:de}.

1) The methods proposed by \citet{vanderwees-bisazza-monz:2017:EMNLP2017} and \citet{zhang-EtAl:2017:I17-2} slightly improved performances. On Test(all), the proposed DSS methods significantly improved the BLEU score by approximately 1.2$\sim$2.2 as compared to the vanilla NMT and by 0.9$\sim$1.7 to the best performing baselines.  As the well-learned sentences increases during NMT training, it did not only slow down NMT training, but also prevent NMT from learning knowledge from the sentences which were not well learned and cause the improvement stagnate. 

2) Within the DSS methods, the review mechanism appears to be a slightly better mechanism than  weighted sampling. This indicates that the review mechanism retained the learned knowledge in a better manner than the learned knowledge of the weighted sampling.

\section{Discussions}
\label{sec:dis}
During the response period, the comments and suggestions of reviewers inspired us a lot. Due to the limited time and space, we briefly discussed these suggestions in this paper. We will show the empirical results in our future work.

\subsection{Effect on Extreme Large Data}
\label{sec:el}
For the large corpus, we have tested the WMT EN-FR task, which containing approximately 12M sentences. The NMT trained from large-scale corpus still gained slight BLEU improvement after several-epoch training.  After 6 epochs training (1M batches), the proposed dynamic sentence sampling method outperformed the baseline by approximately 0.6 BLEU.  

For the web-scale corpora which may be converged within one epoch, in our opinion, if a sentence pair is not well-learned enough, it is necessary to learn it once more. To accelerate this judging processing, we can adopt the sentence similarities between the untrained sentence with small-sized trained sentences as the criteria for sentence sampling. 

\subsection{Effect on Long-time Training}
\label{sec:ltt}
Similarly, for the WMT EN-DE and NIST ZH-EN, if we keep training for more than 1M batches which takes 2-3 weeks, the BLEU would increase by 1.0-1.5 and differences between baseline and the proposed method would slightly decrease by 0.5-0.7 BLEU. Because 7-10 days is a reasonable time for NMT training, we reported 500K batches training results in this paper.

\subsection{Effect on Noisy Data}
\label{sec:nd}

We added 20\% noisy data, which is wrongly aligned, to the NIST ZH-EN corpus. Empirical result shows that the training cost of these noise data did not decrease significantly and even increase sometimes during the training processing. After the first-time time dynamic sampling training by the proposed method, the noise data ratio decreased from 20\% to 13\%. After the second-time dynamic sampling training, the noise data ratio decreased from 13\% to 7\%. This indicates that the proposed method can also detect the noisy data. 

\section{Conclusion}
\label{sec:con}
In this study, the sentences for which training costs of two iterations do not show any significant variation are defined as well-learned sentences. Using a dynamic sentence sampling method, these well-learned sentences are assigned a lower probability of being sampled during the subsequent epoch. The empirical results illustrated that the proposed method can significantly accelerate the NMT training and improve the NMT performances.

\bibliography{long}
\bibliographystyle{acl_natbib}

\end{document}